\documentclass{article}

\usepackage{arxiv}

\usepackage[utf8]{inputenc} 
\usepackage[T1]{fontenc}    
\usepackage{hyperref}       
\usepackage{url}            
\usepackage{booktabs}       
\usepackage{amsfonts}       
\usepackage{nicefrac}       
\usepackage{microtype}      
\usepackage{lipsum}
\usepackage{graphicx}
\usepackage[ruled,vlined]{algorithm2e}
\graphicspath{ {./images/} }
\usepackage{multirow}
\usepackage{array,booktabs}

\newcolumntype{W}{>{\arraybackslash}p{2.1cm}}

\newcolumntype{M}{>{\arraybackslash}p{11cm}}

\title{Sublanguage: A Serious Issue Affects \\Pretrained Models in Legal Domain}

\author{
 Nguyen Ha Thanh \\
  Japan Advanced Institute of Science and Technology\\
  \texttt{nguyenhathanh@jaist.ac.jp} \\
   \And
 Nguyen Le Minh \\
  Japan Advanced Institute of Science and Technology\\
  \texttt{nguyenml@jaist.ac.jp} \\
}

\begin{document}
\maketitle
\begin{abstract}
Legal English is a sublanguage that is important for everyone but not for everyone to understand.
Pretrained models have become best practices among current deep learning approaches for different problems.
It would be a waste or even a danger if these models were applied in practice without knowledge of the sublanguage of the law.
In this paper, we raise the issue and propose a trivial solution by introducing B{\smaller ERT}L{\smaller aw}, a legal sublanguage pretrained model.
The paper's experiments demonstrate the superior effectiveness of the method compared to the baseline pretrained model.

\end{abstract}


\section{Introduction}
Although the ultimate task of NLP is to create models capable of understanding human language, it is a great challenge.
Sometimes, even though we communicate in the same language, we cannot understand each other.
This misunderstanding occurs when the language background knowledge of parties does not match.
In other words, when we do not have enough knowledge about a certain sublanguage, we cannot understand the content represented by it.
For example, in English, languages in medicine, science, and cinema have overlap but are not identical.
Native speakers can have difficulty reading a specialized article if they don't have knowledge of that domain.
Similarly, a deep learning model that works well on the general domain does not guarantee it will work well in a specialized domain.

The formation of sublanguages in a language is inevitable \cite{tiersma1999legal}.
Instead of redefining the concepts every time we exchange ideas, building the specialized vocabulary and syntax in the sublanguage helps us to communicate in a more efficient way \cite{hartig2014plain}.
In communication, humans can naturally invent and learn new concepts by creating and mapping a concept to the corresponding context, thereby forming a semantic model for it.
The language barrier when a lay reader talks to an expert in a particular area is the lack of background knowledge of the sublanguage in that domain.
Unlike other domains, although not being experts in the field of law, people are still bound by rights and obligations by the content written in this sublanguage.

Legal English is a branch of English used in legal writing. 
It has significant differences in terms and linguistic patterns compared to ordinary English.
Legal English is used for writing contracts, regulations, and legal documents that are important for everyone.
The question to ask is whether the models without legal background knowledge can deliver the results that meet real life's expectations.
Our assumption is that if it is difficult for native speakers to understand Legal English, then the deep learning models also face the same problem.
Deep learning models in general and pretrained models in particular are trained based on data.
Hence, unless being familiar with Legal English, they will not work well on tasks that use this sublanguage.

Based on the above points, in this paper, we propose and prove the usefulness of using in-domain data to construct the vocabulary and pretrain a model named B{\smaller ERT}L{\smaller aw}.
Although using the same architecture and training tasks with B{\smaller ERT}, this model gave outstanding results in a legal processing competition.
This is a testament to the point that, if the model architecture is general enough, to build better systems, we need to pay attention to the data.
Before deep learning systems can be involved in part or all of legal work, this is one item of the checklist that needs to be completed.

The main contributions of the paper are as follows.
Firstly, we analyze the problems of legal sublanguage existing in current pretraining techniques. Our assumption is that this issue significantly affects the results of processing legal documents. In this paper, we also refer to other studies about Legal Language and its effect on language understanding.
Secondly, we introduce B{\smaller ERT}L{\smaller aw} as a pretrained model coming along with a specialized vocabulary constructed by data in the legal domain. This model can be a useful resource for law-related downstream tasks.
Thirdly, we conduct experiments to prove the effectiveness of the method. Our approach outperformed using B{\smaller ERT} pretrained on general~data.

\section{Related Work}
\label{sec:related_work}
Although Transformer~\cite{vaswani2017attention} based models are currently the most successful architecture of pretraining methods, this idea predates the advent of these models.
The general idea of pretraining methods is that instead of training the model from scratch, we pretrain the model with a large amount of data so it can learn the main relationships of the concepts in the data.
Once pretrained, the model can perform better on a smaller amount of finetuning data for specific tasks.
In the past decade, a variety of pretraining methods have been introduced and have made constant breakthroughs in NLP.

One of the first pretraining ideas in NLP is using pretrained word embedding.
GloVe \cite{pennington2014glove} and Word2Vec \cite{mikolov2013efficient} are two famous representatives of this approach.
Based on the co-occurrence of the words, the authors create a vector space with the relative position determined by the semantic distance between pairs of words.
Word2Vec uses the prediction model to learn word relationships while GloVe trains a model based on the co-occurrence matrix.
The word vectors represented by these methods not only help to determine the distance between pairs of concepts but also can mathematically interact with each other.
With these vectors, we have equations like "Queen = King - Man + Woman".

ELMo \cite{peters2018deep} approaches the problem from a different perspective from Word2Vec and GloVe.
If the two models above always map a fixed vector value to a corresponding word, the authors of ELMo believe that the word vector can only be determined when there is enough context. Therefore, they propose the concept of contextualized word embedding.
This observation is plausible since homonyms of different meanings exist widely in natural language.
For example, "blue" can be a word for color or a word for mood, we can only determine its exact meaning when there is enough context.
Although this is an important step forward in the pretraining methods of NLP, ELMo's model architecture is bi-directional LSTM~\cite{hochreiter1997long} which is not so excellent in dealing with long contexts.

Improving the shortcomings in earlier works, pretrained models based on Transformer architectures such as B{\smaller ERT} \cite{devlin2018BERT}, BART \cite{lewis2019bart} or GPT-3 \cite{brown2020language} were invented and made a breakthrough in the field of NLP.
Transformer architecture with self-attention and encoder-decoder attention can model the constraints in output and input and represent contextualized word embedding in a longer range. Based on CNN, these models can be computed in parallel, thereby speeding up significantly when training on diverse tasks on huge amounts of data.
Although sharing the same architecture, these models are generated by different pretraining methods on different data.
So far, these models have state-of-the-art performances on benchmark datasets.

The pretraining methods mentioned above all have one thing in common that they use the general domain data.
GloVe is pretrained on Wikipedia dump and Gigaword corpora.
Word2Vec is pretrained on Google News corpus.
ELMo authors used data from Chelba et al. \cite{chelba2013one}, which is a dataset collected and processed from the WMT11 website.
B{\smaller ERT} and BART are trained on BooksCorpus and English Wikipedia.
GPT-3 is trained on Common Crawl, WebText, Book, and Wikipedia datasets.
Therefore, investigating and creating a pretrained model with legal data is essential to achieve high performance in this domain.


\section{Method}
\subsection{The Legal Sublanguage Issue}
Sublanguage is a problem that exists in all domains. Even so, legal sublanguage is a more serious problem and affects many people. We all live and work on the basis of the law. 
The misunderstanding of a law by a person or system can affect the interests of many people, even the whole society. 
As deep learning models increasingly participate in the production and business life of people, this problem is not just a problem for lawyers.
In order to build reliable and effective models, this issue needs to be addressed.

The traditional way of writing by lawyers is considered to be unreadable for a lay reader \cite{butler2013culture}. 
These terms are rarely used in daily conversation but are often used in drafting legal documents such as contracts, terms of service, or statute law.
Legal English critics argue that legal writing should not make a difference in understanding of individuals with different background knowledge \cite{wydick2005plain}. 
However, the actual legal documents still contain vocabulary and structures that are difficult to understand for most people.

Legal English is commonly used in English-speaking countries that share the same legal tradition.
However, this issue is currently not limited to those countries. With widespread economic trade, the language is spoken in a wider range. 
Countries do not translate their laws into English but to Legal English. Besides, international contracts also mainly use this language.
Therefore, the sublanguage of law is becoming a global issue.

This problem is also not limited to human communication but can be a hindrance to deep learning models in NLP.
Pretrained models have become best practices among current deep learning approaches for different problems but if we just use them without realizing the existence of this hindrance, it will lead to great risks.
Hence, investigating and creating pretrained models in this area has scientific and practical significance.

\subsection{B{\smaller ERT} and B{\smaller ERT}L{\smaller aw}}
B{\smaller ERT} is a Transformers-based model that uses a bi-directional context in the training process.
Two unsupervised tasks used to train B{\smaller ERT} are Masked Language Modeling and Next Sentence Prediction.
B{\smaller ERT} also uses subword embedding instead of word embedding or character embedding as in other approaches.
We use the same architecture and the pretraining tasks of B{\smaller ERT} for B{\smaller ERT}L{\smaller aw} but train the model using the legal data. B{\smaller ERT}L{\smaller aw}'s configuration is the same as the original B{\smaller ERT} Base version.

As mentioned in Section \ref{sec:related_work}, since B{\smaller ERT} is trained using the general domain data, we believe that the vocab of this model is not very effective when operating on the legal domain.
B{\smaller ERT} can use subword embedding to technically avoid unknown word problems. For example, 3 words contravention, intervention, and reconvention are tokenized as follows: contra-vent-ion (3 subwords), intervention (1 subword), and rec-on-vent-ion (4 subwords).
Words that do not exist in the vocabulary will be chopped up by B{\smaller ERT} into subwords to represent in its embedding space.
However, the ability to interpret subword correctly needs to be trained with the appropriate context.

We build the vocab for B{\smaller ERT} Law based on our law corpus statistic according to the method of SentencePieces \cite{kudo2018sentencepiece}. 
This approach is language-independent and avoids bias during vocabulary building.
In embedding space, subwords can be considered as unit vectors whose combinations make up the entire space.
The main subwords representing specific relationships in the domain are sampled and added into the vocabulary.

\begin{figure}[h]
  \centering
  \includegraphics[width=.6\linewidth]{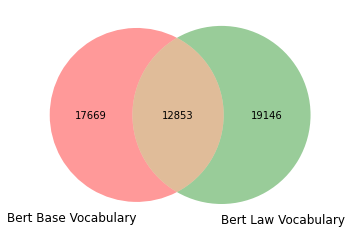}
  \caption{The cardinality of two vocabulary sets of B{\smaller ERT} Base and B{\smaller ERT} Law.}
  \label{fig:venn_diagram}
\end{figure}

Comparing B{\smaller ERT}L{\smaller aw} and B{\smaller ERT} Base vocabulary shows a significant difference in these unit vectors.
Figure \ref{fig:venn_diagram} shows the cardinality of the two vocabulary sets.
The figure shows that the intersection of the two vocabularies is less than half of the total vocabulary for each one.
This shows an obvious difference between the two models in how the same input sentence is observed.

The data we use is the set of American legal cases with 8.2 million sentences (182 million words).
Legal cases contain both vocabularies and grammar in common language and legal language, which enable the model to learn the relationship between the concepts of Legal English and ordinary English.
We pretrain the model with input length 128 tokens using Google Colab's TPU.
The pretraining process ends when the model converges and there is no change in loss value.

Table \ref{tab:examples} shows some examples of subwords that exist in B{\smaller ERT}L{\smaller aw}'s vocabulary but not in B{\smaller ERT}.
From these examples, it can be seen that the embedding space of B{\smaller ERT} is not capable of directly describing the important concepts of the law but indirectly expressed through other subwords. This can affect the performance of the model on downstream tasks.
We provide subwords with explanations of them from the two books "American heritage dictionary of the English language"~\cite{morris1969american} and "A Law Dictionary, Adapted to the Constitution and Laws of the United States"~\cite{bouvier1870law}. Most of these terms are difficult to understand for English users, even for native speakers.

\section{Experiments}

\subsection{Experimental Setting}
We test the model's performance through the Task 4 of COLIEE 2020, an annual competition in automatic legal document processing.
This task is to answer the bar exam questions to assess the competence of paralegal examinees.
Contestants must answer yes/no questions corresponding to each statement given by the exam.
The interesting point in our experiment is that the original data for Task 4 of COLIEE 2020 is Japanese data translated into English by the Ministry of Justice of Japan. This illustrates the globalization of Legal English.

We use the same approach proposed by Nguyen et al.~\cite{jnlp_task4_coliee2019}, consider the task as a lawfulness classification problem. 
Data for creating the model are sentences in civil code, labeled questions provided by the organizer, and the augmented versions of them.
The total data is 5000 samples, of which 90\% is used for training, 10\% for validation.
Test data are provided by the organizer with 112 statements.

The baseline used in our experiment is the original B{\smaller ERT} Base by Google.
We use the max length value for both models as 128.
The models were finetuned on training data in 5 epochs.
The result on the test set is returned by the organizer after the model makes its prediction.

\subsection{Experimental Result}

Table \ref{tab:result} shows the results of the models on validation and test data.
Based on this result, we can see a significant difference between the two models.
B{\smaller ERT}L{\smaller aw} outperforms B{\smaller ERT} Base by almost 4\% on validation data and over 16\% on test data.
This result supports our hypothesis about the effectiveness of pretraining the model on Legal English.

Although 16\% were an impressive result, it was a single-run assessment on the organizers' test data.
The result we care more about is that B{\smaller ERT}L{\smaller aw} outperforms B{\smaller ERT}Base on the validation set consistently.
This shows that the difference in how B{\smaller ERT}L{\smaller aw} and B{\smaller ERT} see the data affects their performance on the downstream task.
Although the unknown word can be overcome by dividing the word into subwords, B{\smaller ERT}'s weight set is not fully trained to model the exact meaning behind those subwords.

\begin{table}
  \caption{B{\smaller ERT}L{\smaller aw} and B{\smaller ERT} Performance on Validation Set and Test Set}
  \label{tab:result}
  \centering
  \begin{tabular}{lcc}
    \toprule
    Model&Validation Accuracy&Test Accuracy\\
    \midrule
    B{\smaller ERT} Base          &0.7784&0.5625\\
    B{\smaller ERT} Law         &0.8168&0.7232\\
  \bottomrule
\end{tabular}
\end{table}

Through the experimental results, we can confirm that pretraining models on the sublanguage of the law help them perform better on tasks in this domain. The experimental data are Japanese civil law and the Japanese legal questions translated into English. B{\smaller ERT}L{\smaller aw} is pretrained using US legal case data. This shows that the terms used in Legal English are consistent across countries. The training of models in this sublanguage is valid not only to English-speaking countries but also to the world.

\begin{table*}
  \caption{Examples appearing in B{\smaller ERT} Law Vocabulary, not in B{\smaller ERT} Base Vocabulary.}
  \label{tab:examples}
  \centering
  \begin{tabular}{|W|M|}
    \hline
    \textbf{Token}& \textbf{Explanation}\\
    \hline
    $\sharp\sharp$legal&Wordpiece in words containing ``legal" (e.g. illegal, legally, legality, legalization)\\\hline
    
    contravention&An act which violates the law, a treaty or an agreement which the party has made. \cite{bouvier1870law}\\\hline
    
    construe&To determine the meaning of the words of a written document, statute or legal decision, based upon rules of legal interpretation as well as normal meanings. \cite{bouvier1870law}\\\hline

    demurrer&An assertion by the defendant that although the facts alleged by the plaintiff in the complaint may be true, they do not entitle the plaintiff to prevail in the lawsuit. \cite{bouvier1870law}\\\hline
    
    depose&To make a deposition; to give evidence in the shape of a deposition; to make statements that are written down and sworn to; to give testimony that is reduced to writing by a duly qualified officer and sworn to by the deponent. \cite{bouvier1870law}\\\hline

    guardianship& The power or protective authority given by law, and imposed on an individual who is free and in the enjoyment of his rights, over one whose weakness on account of his age, renders him unable to protect himself. \cite{bouvier1870law}\\\hline
    
    infringe&To transgress or exceed the limits of; violate: infringe a contract; infringe a patent. \cite{morris1969american}\\\hline
    
    malfeasance&The commission of an act that is unequivocally illegal or completely wrongful. \cite{bouvier1870law}\\\hline
    
    misdemeanor&Offenses lower than felonies and generally those punishable by fine, penalty, forfeiture, or imprisonment other than in a penitentiary. \cite{morris1969american}\\\hline
    
    reimburse&To repay (money spent); refund. \cite{morris1969american}\\\hline
    
    renounce& To give up a right; for example, an executor may renounce the right of administering the estate of the testator; a widow the right to administer to her intestate husband's estate. \cite{morris1969american}\\\hline
    
    rescind& To declare a contract void—of no legal force or binding effect—from its inception and thereby restore the parties to the positions they would have occupied had no contract ever been made. \cite{morris1969american}\\\hline
    
    rescission&The termination of a contract by mutual agreement or as a result of fraud or some legal defect. \cite{morris1969american}\\\hline
    
    revoke&To invalidate or cause to no longer be in effect, as by voiding or canceling. \cite{morris1969american}\\\hline

    tort& a civil wrong. Tortious liability arises from the breach of a duty fixed by law; this duty is towards persons generally and its breach is redressable by an action for unliquidated damages. \cite{morris1969american}\\\hline
    tortious&Wrongful; conduct of such character as to subject the actor to civil liability under Tort Law. \cite{morris1969american}\\

  \hline
\end{tabular}
\end{table*}
\section{Discussions}
Although B{\smaller ERT}L{\smaller aw} is a concrete instance of this approach, it is possible to create pretrained models based on other architectures and other tasks using this approach.
The important message of this paper is that using only the pretrained models as a tool without examining the characteristics of the pretraining data will not fulfill their full potential.
In addition, along with benchmark datasets for general domains, we also need benchmark datasets for specialized areas in order to better verify different solutions to sublanguage problems.
Without data, all proposals are only theoretical and difficult to apply in practice.

In our opinion, among different domains, law is the one that should be given priority in solving the sublanguage problem.
As repeated many times in this paper, law is an area that concerns all citizens of every country and everyone.
Solving this problem helps us move closer to our goal of having safe, reliable, and equal AI systems.
Besides, governments are building policies to regulate AI systems and they need to be accountable.
We believe that in the near future, these systems will not be able to explain their decisions to the legal agency without understanding the correct language of the law.

\section{Conclusions}
In this study, we address the legal sublanguage issue for the performance of deep learning systems.
Our assumption is that if people have difficulty in understanding Legal English, this is also a barrier to deep learning systems.
To verify our assumptions, we created a pretrained model called B{\smaller ERT}L{\smaller aw} and compared its performance with original B{\smaller ERT} Base from Google.
The experimental results support our hypothesis.
In addition, we discuss the reasons for those results as well as the impacts of this finding.
B{\smaller ERT}L{\smaller aw} and this paper are useful references and resources for legal domain problems to be solved by deep learning.

\bibliographystyle{plain}      
\bibliography{./references}   

\end{document}